\documentclass[runningheads]{llncs}
\usepackage[T1]{fontenc}
\usepackage[utf8]{inputenc}
\usepackage{amsmath,amssymb}
\usepackage{booktabs}
\usepackage{multirow}
\usepackage{xcolor}
\usepackage{graphicx}
\usepackage{algorithm}
\usepackage{algpseudocode}
\usepackage[hidelinks]{hyperref}
\usepackage{xspace}
\newif\ifshowtodos \showtodosfalse

\newcommand{\REPOURL}{\url{https://github.com/SniperJF/ocr-edit-gate-release}}
\newcommand{\HFURL}{\url{https://huggingface.co/SJF7}}
\newcommand{\cmer}{\mathrm{cMER}}
\newcommand{\pref}{\mathrm{Pref}}
\newcommand{\norm}{\nu}
\newcommand{\sysname}{\textsc{EditVeto}\xspace} 
\newcommand{\hipe}{HIPE-OCRepair-2026\xspace}

\DeclareMathOperator*{\argmax}{arg\,max}

\makeatletter
\let\egate@fsruled\fs@ruled
\renewcommand\fs@ruled{\egate@fsruled\def\@fs@cfont{\color{black}\bfseries}}
\makeatother

\newcommand{\apptabref}[1]{\iforacleappendix\ifextended Table~\ref{#1} in the appendix\else the extended version of this paper\fi\else Table~\ref{#1}\fi}
\newcommand{\appref}[1]{\ifextended Appendix~\ref{#1}\else the extended version of this paper\fi}

\newcommand{\gatebb}{\mbox{Qwen3-0.6B}\xspace}

\newif\ifanonymous \anonymousfalse
\newif\iforacleappendix \oracleappendixtrue
\newif\ifextended \extendedtrue
\newif\ifarxiv \arxivtrue

\begin{document}

\title{Rewrite, Then Veto: Rewriter-Agnostic Edit Gating for LLM-Based Post-OCR Correction}
\titlerunning{Rewriter-Agnostic Edit Gating for Post-OCR Correction}

\ifanonymous
  \author{Anonymous Authors}
  \authorrunning{Anonymous}
  \institute{Anonymous Institution}
\else
  \author{Jorge Ram\'on Fonseca Cacho\inst{1}\orcidID{0000-0001-7917-5169}}
  \authorrunning{J. R. Fonseca Cacho}
  \institute{Department of Computer Science, University of Nevada, Las Vegas, NV, USA\\
  \email{jorge.fonsecacacho@unlv.edu}}
\fi

\maketitle

\begin{abstract}
LLMs can repair noisy OCR transcripts of historical documents, but they also change text that was already correct. We propose \sysname, a veto gate placed after a full-context LLM rewrite. The rewrite is character-aligned to the OCR input, its differences are split into blocks, and a 0.6B-parameter gate keeps or reverts each block. An oracle analysis motivates the design: given the true error locations, per-span generation is worse than no correction, whereas vetoing a full rewrite's edits helps. Trained on rewrites by two Qwen3 models and evaluated, frozen, on the eight \hipe test sets, the gate improves all five rewriters tested, two of them unseen in training: weighted cMER falls by 0.0026--0.0044 and the preference score rises by 0.15--0.25 (paired bootstrap, $p\approx0.0002$ each). On published competition outputs it lowers cMER for three of four teams but raises it for the winning run (0.0050 to 0.0065). In a second, disclosed test look, a retrained gate still improves LoRA-tuned rewriters (cMER $-0.0011$ to $-0.0021$), and an ensemble of their gated edits reaches a cMER of 0.0069 and a preference score of 0.928, mostly through fine-tuning and pooling rather than gating. Built only on public models and data, it is behind the winner (0.0050) and tied with the second-ranked run (0.0071), both from the only team using domain pre-training, and ahead of all other published runs. A third, secondary look shows the unchanged gate improving zero-shot \mbox{Qwen3.8-27B} (cMER 0.0114 to 0.0108) but only the preference score of its LoRA-tuned version.
\end{abstract}

\section{Introduction}
\label{sec:intro}

Libraries and archives hold large collections of historical newspapers and books whose text layer was produced by optical character recognition (OCR) years or decades ago. Re-digitizing these collections is rarely feasible, so their usefulness for search and text analysis depends on post-correction: improving an existing transcript without access to the page image~\cite{nguyen2021survey,ehrmann2026hipe}. Post-correction has moved from noisy-channel models~\cite{kolak2002noisy} and neural sequence-to-sequence systems~\cite{amrhein2018supervised,soper2021bart,rijhwani2020endangered} to LLMs used zero-shot or after fine-tuning~\cite{boros2024postcorrection,thomas2024leveraging,kanerva2025nofreelunches,bourne2025scrambled}.

Post-correction is a constrained task: a system should restore the intended word forms and must not add, remove, or modernize content~\cite{ehrmann2026hipe}. LLMs do not respect this constraint by default. The \hipe competition at ICDAR~2026~\cite{ehrmann2026hipe} released a multilingual benchmark (English, French, German; 17th--20th century) scored by micro-averaged character match error rate (cMER) and a per-unit preference score that counts how often a system improves rather than degrades a unit. The organizers report over-correction of low-noise inputs as a recurring challenge and list methods to control over-correction and hallucination as future work~\cite{ehrmann2026hipe}. The winning system adapted a 24B-parameter model with continued pre-training on roughly 80B tokens, fine-tuning, and multi-step inference, and reached a cMER of 0.0050 and a preference score of 0.9000 on the official test sets~\cite{ehrmann2026hipe}. Such adaptation is likely beyond the compute budget of many libraries and archives.

We ask whether over-correction can be controlled \emph{after} generation, independently of the model that corrects. Our starting point is a negative result. Given the exact error locations from the ground truth, a \mbox{Qwen3-8B} model that generates a replacement for each marked span is worse than no correction on the development set (cMER 0.0562 against 0.0275). By contrast, letting the same model rewrite the whole unit and then reverting every edit block that does not touch an oracle error span improves both metrics over the unfiltered rewrite. For this model and prompt, the useful role of an edit gate is therefore not to drive generation span by span but to \emph{veto} the collateral edits of a full-context rewrite.

We build on this observation with \sysname. A rewriter proposes a corrected unit. We align the rewrite to the OCR input at the character level, split their differences into edit blocks, and let a small learned gate accept or veto each block. A vetoed block is restored to the OCR text, so the output contains only text that appears in the input or in the rewrite. The gate is trained once, on blocks from two Qwen3 rewriters on the training split, and then frozen. We call it rewriter-agnostic in a narrow sense: its text-only variant needs nothing from the rewriter but the output text, and we apply it without retraining to rewriters and published system outputs it never saw. It does not help every system; Sect.~\ref{sec:test} shows where it hurts. In a second, disclosed test evaluation we combine a retrained gate with LoRA-tuned rewriters and an ensemble that pools their gated edits. This shows how far the approach goes with stronger rewriters, but it changes what is evaluated: the gains there come mostly from fine-tuning and pooling, and we report the gate's own share separately. A third, secondary look applies the same gate, unchanged, to a newer rewriter. Our contributions are:
\begin{itemize}
  \item A formulation of post-correction as edit vetoing over an optimal character alignment, with an exact characterization of the vetoed output on raw strings and its consequences under the official normalized scorer (Sect.~\ref{sec:problem}).
  \item An oracle analysis that separates knowing \emph{where} to edit from the rewriter's ability to edit, with two negative results for per-span generation (Sect.~\ref{sec:oracle}).
  \item A block-level veto gate with a \gatebb backbone, trained and selected on the training split only, in a full and a text-only variant, that transfers without retraining to rewriters outside its training pair (Sect.~\ref{sec:results}).
  \item A first frozen evaluation on the eight \hipe test sets: the gate improves all five rewriters on both metrics, and lowers cMER for every run of three of the four competition teams but raises it for two runs of the lowest-cMER team (Sect.~\ref{sec:test}).
  \item A second, disclosed test evaluation of LoRA-tuned rewriters with a retrained gate and an edit-pool ensemble, which separates the gate's effect from that of fine-tuning and pooling, and two additions that gave no gated gain on development data, sampled candidates and synthetic training pairs, and a secondary third look with \mbox{Qwen3.8-27B} (Sect.~\ref{sec:phase2}).
\end{itemize}

\section{Related Work}
\label{sec:related}

\paragraph{Post-OCR correction.}
Nguyen et al.~\cite{nguyen2021survey} survey post-OCR processing. Early systems decoded in a noisy channel~\cite{kolak2002noisy}, and context was shown to outweigh confusion statistics for noisy newspaper OCR~\cite{evershed2014correcting}. Neural systems treated correction as translation from OCR to clean text~\cite{amrhein2018supervised,rijhwani2020endangered,soper2021bart}. Schaefer and Neudecker~\cite{schaefer2020twostep} separated detection from correction and reported fewer falsely changed correct characters; our gate also separates the two decisions, but after correction, on the edits a rewriter has proposed. Alignment of OCR output with ground truth underlies training and error analysis~\cite{fonsecacacho2019aligning}, candidates have been generated with BERT and subword embeddings~\cite{hajiali2021generating}, and the ICDAR 2017 and 2019 competitions set shared evaluations~\cite{chiron2017icdar,rigaud2019icdar}.

\paragraph{LLM-based post-OCR correction.}
Boros et al.~\cite{boros2024postcorrection} evaluated fourteen foundation models in zero- and few-shot settings and concluded that LLMs were not yet effective at the task. Thomas et al.~\cite{thomas2024leveraging} instruction-tuned Llama~2 on 19th-century British newspapers and reported a larger character error reduction than a fine-tuned BART, and Kanerva et al.~\cite{kanerva2025nofreelunches} found useful gains for English but not for Finnish. Bourne~\cite{bourne2025scrambled} found that fine-tuning on synthetic data from a character-level Markov corruption process can outperform real data. Thomas et al.~\cite{thomas2026image} add the page image through vision-language models; we stay in the text-only setting of \hipe~\cite{ehrmann2026hipe}. Sastre et al.~\cite{sastre2025constrained} constrain the rewriter itself, interpolating its next-token distribution during beam search with a character-similarity distribution to the OCR, and report fewer lines made worse; this needs access to the decoder. Our gate leaves decoding unchanged, makes discrete decisions per edit block, and needs only input and output text.

\paragraph{Edit-based and constrained generation.}
Tagging-based grammatical error correction predicts per-token edit operations and keeps tokens by default~\cite{malmi2019encode,omelianchuk2020gector}, and recent work adapts decoder-only LLMs to minimal-edit correction~\cite{staruch2025minimal}. Copy mechanisms let sequence-to-sequence models reproduce input tokens~\cite{gu2016copying}, and lexically constrained decoding forces specified phrases into the output~\cite{hokamp2017lexically}. These methods share our premise that most of the input should be preserved, but they shape generation. \sysname instead filters a finished rewrite at the level of character edit blocks, which matches the form of OCR noise: character substitutions, insertions, and deletions together with layout artifacts~\cite{ehrmann2026hipe}.

\section{Problem Definition}
\label{sec:problem}

\paragraph{Task.}
Following \hipe~\cite{ehrmann2026hipe}, a \emph{transcription unit} (a paragraph, article, page, or chunk) consists of an OCR hypothesis $x \in \Sigma^{*}$ over a character alphabet $\Sigma$ and metadata $m$ (language and, where available, approximate date and publication title). The system has no access to the page image and outputs a corrected transcription $\hat{y}$, compared with a manually verified reference $y^{\star}$.

\paragraph{Metrics.}
Before scoring, the official scorer applies a normalization $\norm(\cdot)$ to both output and reference~\cite{ehrmann2026hipe}. In the scorer code (version 0.9.9),\footnote{\url{https://github.com/hipe-eval/HIPE-OCRepair-scorer}, function \texttt{norm}.} $\norm$ lowercases the text; maps a few historical characters (e.g.\ \textit{\ss}$\to$\textit{ss}, r~rotunda$\to$\textit{r}); deletes the hyphenation markers \textit{¬} and \textit{---} before a line break, joining the word parts; replaces every non-word character by a space; and collapses and strips whitespace. For a character-level alignment between $\norm(\hat{y})$ and $\norm(y^{\star})$ with $H$ hits, $S$ substitutions, $D$ deletions, and $I$ insertions, the character match error rate is $\cmer=(S+D+I)/(H+S+D+I)\in[0,1]$. The primary criterion is the micro average, which sums $H,S,D,I$ over all units of a test set before taking the ratio~\cite{ehrmann2026hipe}. The secondary criterion is the preference score. Let $e_i^{\mathrm{ocr}}$ and $e_i^{\mathrm{cor}}$ be the cMER of the OCR input and of the system output for unit $i$; then $\pref=\frac{1}{N}\sum_{i=1}^{N}\operatorname{sign}(e_i^{\mathrm{ocr}}-e_i^{\mathrm{cor}})\in[-1,1]$, so a unit counts $+1$ if improved, $0$ if unchanged in error, and $-1$ if degraded~\cite{ehrmann2026hipe}. We use \emph{over-correction} for edits to text that was already correct. Even the noisiest ranked test set has an average per-unit (macro) test CER after normalization of only 0.057 (\texttt{dta19-l2}; 0.004 on \texttt{dta19-l0})~\cite{ehrmann2026hipe}, so most characters of most units are correct, and a system that edits freely risks degrading many units in exchange for gains on a few. The preference score exposes this trade-off.

\paragraph{Edit vetoing.}
A \emph{rewriter} $\mathcal{R}$, any system that maps $(x,m)$ to a full corrected unit $z=\mathcal{R}(x,m)$, proposes a correction. Let $d(\cdot,\cdot)$ be the Levenshtein distance~\cite{levenshtein1966binary} and let $\Omega$ be a minimum-cost edit script of unit-cost substitutions, insertions, and deletions that turns $x$ into $z$, so $|\Omega|=d(x,z)$. Grouping maximal runs of operations of $\Omega$ that are not separated by a matched character into \emph{edit blocks} gives
\begin{equation}
  x = u_0\,B_1\,u_1 \cdots B_n\,u_n, \qquad z = u_0\,B'_1\,u_1 \cdots B'_n\,u_n,
  \label{eq:decomposition}
\end{equation}
where the $u_i$ are the aligned common segments and block $i$ rewrites $B_i$ into $B'_i$ (either may be empty). A veto gate chooses a set $A \subseteq \{1,\dots,n\}$ of accepted blocks, and the output is $\hat{y} = u_0\,c_1\,u_1 \cdots c_n\,u_n$ with $c_i = B'_i$ for $i \in A$ and $c_i = B_i$ (a vetoed block, restored to the OCR text) otherwise. Every character of $\hat{y}$ thus comes from $x$ or from $z$.

\paragraph{A damage bound.}
Let $E(w)=S+D+I$ be the error count the scorer assigns to an output $w$ against $y^{\star}$. The scorer obtains $S$, $D$, and $I$ from a minimum-cost alignment of $\norm(w)$ with $\norm(y^{\star})$,\footnote{It calls \texttt{jiwer.process\_characters}, which takes the edit operations from \texttt{rapidfuzz.distance.Levenshtein.opcodes}.} so $E(w)=d(\norm(w),\norm(y^{\star}))$.

\smallskip\noindent\textbf{Proposition 1.} \emph{For any rewrite $z$ and accepted set $A$: (i) if $A=\emptyset$, then $\hat{y}=x$, and the unit's cMER and preference contribution equal those of no correction; (ii) $d(x,\hat{y})=\sum_{i\in A} d(B_i,B'_i)$ and $d(\hat{y},z)=\sum_{i\notin A} d(B_i,B'_i)$, so $\hat{y}$ lies on a shortest edit path from $x$ to $z$; (iii) under the official scorer,}
\begin{equation}
  E(\hat{y}) \;\le\; \min\Bigl\{\,E(x) + d\bigl(\norm(\hat{y}),\norm(x)\bigr),\;\; E(z) + d\bigl(\norm(\hat{y}),\norm(z)\bigr)\Bigr\}.
  \label{eq:bound}
\end{equation}

\noindent The proof is short: the operations of $\Omega$ inside each block form a minimum-cost script for that block, so the accepted and vetoed parts split $d(x,z)$ exactly, and (iii) follows from the triangle inequality on normalized strings (\appref{app:proof}). Statement (ii) concerns raw strings; its normalized analogue can fail, because $\norm$ joins hyphenated words across line breaks. Part (iii) holds for any output; its use here is that the veto controls the distance terms. On all 4,488 logged development and probe outputs, the script returned by \texttt{rapidfuzz} \texttt{Levenshtein.editops} has minimum length, and (i)--(iii) hold for every tested block subset (22 per output); the normalized analogue of (ii) fails for 1,375 of the 98,736 subsets.

\section{Method}
\label{sec:method}

\begin{figure}[t]
\centering
\includegraphics{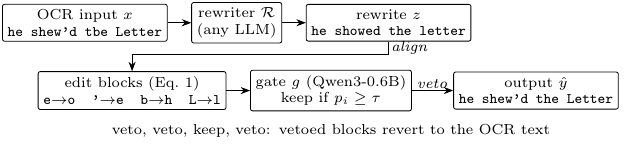}
\caption{Overview of \sysname on a constructed example (not from the data). The rewrite is aligned to the OCR input, each edit block is kept or vetoed, and vetoed blocks revert to the OCR text. Here the gate keeps the OCR fix and vetoes the modernized spelling and the case change.}
\label{fig:overview}
\end{figure}

\sysname has three parts (Fig.~\ref{fig:overview}): a rewriter that proposes a full corrected unit (Sect.~\ref{sec:rewriter}), an alignment step that turns the proposal into edit blocks (Sect.~\ref{sec:alignment}), and a gate that accepts or vetoes each block (Sect.~\ref{sec:gate}).

\subsection{Rewriters}
\label{sec:rewriter}
The rewriter sees the whole transcription unit, so it can use document-level context. The alignment and the veto need only its output text; the full gate also uses its token log-probabilities (Sect.~\ref{sec:gate}). Our rewriters are instruction-tuned LLMs prompted zero-shot to correct OCR errors, keeping historical spelling, wording, punctuation, and line breaks, without modernizing, paraphrasing, or completing truncated text, and LoRA-adapted~\cite{hu2021lora} models (Sects.~\ref{sec:systems} and~\ref{sec:phase2}).

\subsection{Alignment and Edit Blocks}
\label{sec:alignment}
We compute a minimum-cost edit script under unit costs between $x$ and $z$ with the Levenshtein dynamic program~\cite{wagner1974string,levenshtein1966binary}, as implemented by \texttt{rapidfuzz} (\texttt{Levenshtein.editops}), and merge adjacent non-matching operations into edit blocks, as in Eq.~\eqref{eq:decomposition}. A block is accepted or vetoed as a whole (the oracle analysis also decides single operations); among equal-cost scripts, the aligner's choice sets the block boundaries.

\paragraph{Block labels and reference-based gates.}
For analysis and training we label every block with the reference. Let $x\oplus i$ be the OCR text with only block $i$ applied. The label of block $i$ is the sign of $g_i=\cmer(x)-\cmer(x\oplus i)$, computed per unit with the scorer's own functions on normalized text: \emph{helpful} if positive, \emph{harmful} if negative, and \emph{neutral} if zero (e.g.\ edits of case or punctuation only, which the scorer ignores). Because $\norm$ couples neighboring characters, single-block effects are not exactly additive. Three reference-based gates serve as points of comparison; none is a valid system. The \emph{oracle span gate} aligns $x$ with $y^{\star}$, grows each non-matching operation to the boundaries of its whitespace-delimited token, merges spans separated only by whitespace, drops spans whose reference replacement leaves $\norm(x)$ unchanged, and accepts a block if it overlaps a span (an insertion-only block if it touches one). It is the gate of Sect.~\ref{sec:oracle}. The \emph{label oracle} accepts exactly the helpful blocks (not a strict upper bound, as block effects are not additive). The \emph{reverse} span gate accepts only the blocks the oracle span gate vetoes.

\subsection{Learned Veto Gate}
\label{sec:gate}
\paragraph{Model and features.}
For each block $i$ the gate outputs a score $p_i=g(x,z,i)$, the estimated probability that the block is helpful. Its text input is a language and dataset tag, the OCR context of $\pm$60 characters with the block marked, and the proposed replacement, truncated to 256 tokens. A \gatebb backbone~\cite{qwen3} encodes it with last-token pooling. The pooled vector is concatenated with 15 standardized numeric features and fed to an MLP head (one hidden layer of 256 units, GELU, dropout 0.1). Six features are statistics of the rewriter's token log-probabilities over the block (mean, minimum, sum, count, first token, unit mean), and one flags a failed token mapping. The other eight need only the two texts (block lengths and their difference, log unit length, log block count, fraction changed, insertion and deletion flags). The \emph{text-only} gate has the same architecture and training but these seven log-probability features (the flag included) set to zero, so it applies to any rewriter's output, including published submission files.

\paragraph{Training data.}
Both frozen gates are trained on the pooled blocks of zero-shot \mbox{Qwen3-8B} and \mbox{Qwen3-32B} rewrites of the full training split (2,402 units; 89,602 and 167,531 blocks, of which 57,459 and 102,931 are helpful or harmful). Neutral blocks are excluded, as are three training units whose reference also occurs in the development set.\ifanonymous\else{} Both gates are released (Sect.~\ref{sec:availability}).\fi\ For selection we hold out, per cell and with seed 1234, 15\% of the groups of training units that share a reference text. Because each \texttt{dta19} page appears at three noise levels, 21\% of the units are held out (466 of 2,179 and 498 of 2,371). This leaves 130,165 labeled blocks for fitting and 30,077 held out. Training uses unweighted binary cross-entropy, AdamW (weight decay 0.01; learning rates $3\cdot10^{-5}$ backbone, $10^{-3}$ head; 5\% warm-up, linear decay), batch size 32, 3 epochs, bf16, seed 1234.

\paragraph{Selection and decision rule.}
The scores are used without post-hoc calibration. The gate keeps block $i$ if $p_i\ge\tau$; a rejected block reverts to the OCR text. On the held-out training blocks we choose the epoch and one global threshold that maximize the total labeled gain of the kept blocks, $\tau=\argmax_t\sum_{i:\,p_i\ge t} g_i$, which is the sum of the per-unit cMER reductions the scorer would see if block effects were additive. For both pooled gates this gives epoch~1, with $\tau=0.5065$ for the full gate and $\tau=0.4990$ for the text-only gate. Among the \gatebb gates we froze the two with the highest held-out ROC-AUC (0.952 and 0.949); no development number entered this choice, and there is no per-language, per-cell, or per-rewriter setting. As training-free baselines we threshold the mean or the minimum token log-probability of the block, with the threshold chosen on the fit part of the training data.

\section{Experimental Setup}
\label{sec:setup}

\subsection{Benchmark and Splits}
\label{sec:benchmark}

We use the \hipe dataset~\cite{ehrmann2026hipe}, release v0.9 (tag v0.9.5). It consolidates parallel OCR/reference pairs from historical newspapers and printed books in English, French, and German (17th--20th century). The official ranking uses eight test sets~\cite{ehrmann2026hipe}: \texttt{dta19} at three OCR noise levels (German Fraktur books; \texttt{l1} and \texttt{l2} are OCR of page images degraded to target about 3\% and 7\% CER), \texttt{icdar2017} (English and French newspapers), and \texttt{impresso-snippets} (German, English, and French newspaper paragraphs). The released test references hold 740 units, 80 per \texttt{dta19} level, but the official scorer drops the 50 units per level that are flagged \texttt{exclude\_from\_icdar\_evaluation}; 590 units are scored, 30 per \texttt{dta19} level, which matches the counts in Table~2 of~\cite{ehrmann2026hipe}. Development and test references were manually revised; training references were not~\cite{ehrmann2026hipe}. The data are used under their licenses (CC~BY-NC-SA~4.0, CC~BY-NC~4.0, CC~BY-SA~4.0, and research-only use for \texttt{overproof})~\cite{ehrmann2026hipe}; we do not redistribute them. We use \texttt{impresso-nzz} (CC~BY-NC~4.0) for non-commercial research only. We do not redistribute the synthetic pairs or the adapter trained on them, as their sources include share-alike and non-commercial text.

\paragraph{Development set.} All design decisions use the training and development splits. The development split has 518 units in seven of the eight official cells, plus 30 \texttt{overproof} units (548 in total): \texttt{icdar2017}-en (188 units), \texttt{impresso-snippets} en/fr/de (10 each), and the three \texttt{dta19} levels (100 each, the \emph{unmatched} files, as the official DTA test sets are unmatched). \texttt{icdar2017}-fr has no development file~\cite{ehrmann2026hipe}, so development means cover 7 of 8 cells and are not directly comparable to test means. The \texttt{overproof} file is scored but excluded from the means, as it is not an official cell.

\subsection{Metrics, Statistics, and Protocol}

We score all systems with the official scorer (\texttt{score\_one.py} of the evaluation repository,\footnote{\url{https://github.com/hipe-eval/HIPE-OCRepair-2026-eval}; we use commit \texttt{47a71d8}.} which calls \texttt{hipe-ocrepair-scorer} 0.9.9). For each cell we report micro cMER (primary) and the preference score (secondary). Overall numbers are the official weighted mean of per-cell values: each \texttt{dta19} level has weight~1/3 and every other cell weight~1~\cite{ehrmann2026hipe}; development means use the same weights over seven cells. Recomputed with this scorer, all 58 published submission files reproduce the organizers' per-cell and overall values to four decimals.

\paragraph{Paired bootstrap.} We recompute per-unit counts with the scorer's own functions, check them against the scorer's per-cell output, and resample units with replacement independently within each cell, using the same indices for both systems (10,000 resamples, seed 1234). The weighted difference is recomputed on each resample; we report its 2.5th and 97.5th percentiles as a 95\% CI and a two-sided $p$ with the $+1$ correction, whose smallest value is $2/10{,}001\approx0.0002$. Tables show the scorer's values, which can differ by 0.0001 from bootstrap means.

\paragraph{Test looks.} We scored the test split more than once, and each look is reported. \emph{Look~1} (Sect.~\ref{sec:test}): the gates, thresholds, prompt, decoding parameters, and rewriter list were frozen after the development experiments, with checksums of the gate files; before it, only a no-correction sanity check had been scored on test. It was run once, with no later change to look~1 itself (look~2 regenerated one look-1 rewriter's outputs), and also gated the published competition outputs with the text-only gate. Three harness defects were found by inspection before the affected step ran and fixed without changing any frozen quantity (\appref{app:repro}). \emph{Looks~2 and~3} (Sect.~\ref{sec:phase2}) evaluated later systems, look~3 only pre-declared secondary ones; each was run once, with no harness change. \ifanonymous Code (GitHub) and the trained gates and adapters (Hugging Face; except LoRA-72B-syn) are released; links are withheld for double-blind review.\else Code, gate configurations, checksums, and run logs are released at \REPOURL; the trained gates and adapters (except LoRA-72B-syn) are on Hugging Face (\HFURL); see Sect.~\ref{sec:availability}.\fi

\paragraph{Selection protocol and overfitting.} Gates, thresholds, epochs, and pools were selected on held-out training data (two steps also saw development scores), each configuration was hash-frozen before its look, and all three looks are reported, with ensemble~B declared primary before look~2. The held-out ranking of ensembles B and C (Sect.~\ref{sec:phase2}) carried over to test (0.0076 and 0.0073 held-out; 0.0069 and 0.0066 test). Repeated looks still risk adaptive overfitting, and later work followed look~1 (e.g.\ where the gate hurt), so later looks deserve more caution (\appref{app:look2}).

\subsection{Rewriters and Baselines}
\label{sec:systems}
\label{sec:baselines}

\textbf{Rewriters.} All rewriters receive the same zero-shot prompt and the whole unit and return the whole corrected unit, with greedy decoding (seed 1234), at most $\max(512, 0.6\,|x|)$ new tokens for an input of $|x|$ characters, and a fallback to the OCR text for empty, truncated, or failed outputs. They are \mbox{Qwen3-8B} and \mbox{Qwen3-32B}~\cite{qwen3} with thinking mode off, whose training-split rewrites train the gate; \mbox{Mistral-Small-24B}-Instruct-2501\footnote{\url{https://huggingface.co/mistralai/Mistral-Small-24B-Instruct-2501}} and \mbox{Qwen2.5-72B}-Instruct~\cite{qwen25}, whose outputs the gate never saw in training; and \mbox{Qwen3-8B} with a LoRA adapter~\cite{hu2021lora} (rank 16, $\alpha=32$, one epoch, learning rate $10^{-4}$) trained on 2,250 training units, without \texttt{impresso-nzz}\ifanonymous\else{} (released; Sect.~\ref{sec:availability})\fi. Because the LoRA rewriter was tuned on the training split, its training rewrites were not used to train a gate. Models are served with vLLM 0.32.0 on two NVIDIA H200 GPUs. The development work of Sect.~\ref{sec:oracle} used 1.08 GPU-hours, the gate and rewriter development 5.25, and the test run 1.10. \textbf{Oracle per-span systems} (Sect.~\ref{sec:oracle}) give \mbox{Qwen3-8B} the oracle spans: per-span generation sees about 300 characters of context per side and returns a replacement; the window rewrite rewrites up to 40 characters per side and keeps only edits inside the span. The learned-gate experiments use a new \mbox{Qwen3-8B} run with token log-probabilities, which differs from the run of Sect.~\ref{sec:oracle} for 89 of 548 development units (0.0245 / 0.4610 against 0.0247 / 0.4161).

\textbf{Baselines.} No correction returns the OCR text and has a preference score of zero. The competition systems, as described by the organizers~\cite{ehrmann2026hipe}, are \textbf{BnF-Mistral} (runs 1--3; run~1 ranked first), which uses Mistral Small~3 (24B) after continued pre-training on about 80B tokens, LoRA fine-tuning on competition and synthetic data, a judge-and-retry loop, and routing of difficult documents to a larger model; \textbf{BLOCR}, which prompts DeepSeek (English) and Gemma~3 (French, German) zero-shot; \textbf{L3i}, which instruction-tunes \mbox{Qwen3-14B} (run~2 covers three cells); and \textbf{Zakaria-ENSIAS}, which prompts \mbox{Qwen3-8B} zero-shot with a length-ratio safeguard and did not submit \texttt{dta19-l2}.

\section{Motivation: Oracle Analysis (Development Set)}
\label{sec:oracle}

Before training a gate we asked, with reference-based gates on the development set, whether knowing \emph{where} to edit helps and through which interface (details in \apptabref{tab:oracle}), using the earlier \mbox{Qwen3-8B} zero-shot run (0.0247 / 0.4161; Sect.~\ref{sec:systems}). These gates and generators read the reference and are not valid systems. Means exclude \texttt{overproof}; the span counts below include it.

\iforacleappendix\else\begin{table}[t]
\centering
\caption{Development set, \emph{oracle} analysis (earlier \mbox{Qwen3-8B} zero-shot run; see Sect.~\ref{sec:systems}): weighted means over the seven official development cells (518 units). Oracle systems read error locations from the reference and are not valid systems.}
\label{tab:oracle}
\small\setlength{\tabcolsep}{5pt}
\begin{tabular}{lcc}
\toprule
System & cMER $\downarrow$ & Pref $\uparrow$ \\
\midrule
No correction & 0.0275 & 0.0000 \\
Zero-shot rewrite (ZS) & 0.0247 & 0.4161 \\
\midrule
Oracle per-span generation (earlier design) & 0.0562 & $-$0.6247 \\
Oracle window rewrite & 0.0255 & 0.2627 \\
\midrule
ZS + oracle span gate (block) & \textbf{0.0229} & \textbf{0.6501} \\
ZS + oracle span gate (character) & 0.0236 & 0.5681 \\
ZS + reverse span gate (block; sanity check) & 0.0292 & $-$0.3181 \\
\bottomrule
\end{tabular}
\end{table}
\fi

\iforacleappendix\else\begin{table}[t]
\centering
\caption{Development set, \emph{oracle} analysis: per-cell micro cMER and preference score (earlier \mbox{Qwen3-8B} zero-shot run). ZS+OSG: zero-shot rewrite with the oracle span gate (block). OPG: oracle per-span generation. OW: oracle window rewrite. Units per cell in parentheses.}
\label{tab:oraclecell}
\scriptsize\setlength{\tabcolsep}{2.6pt}
\begin{tabular}{lcccccccc}
\toprule
 & No corr. & \multicolumn{2}{c}{ZS} & \multicolumn{2}{c}{ZS + OSG} & OPG & \multicolumn{2}{c}{OW} \\
\cmidrule(lr){3-4}\cmidrule(lr){5-6}\cmidrule(lr){8-9}
Cell & cMER & cMER & Pref & cMER & Pref & cMER & cMER & Pref \\
\midrule
\texttt{icdar2017}-en (188)  & 0.0343 & 0.0288 & 0.6436 & 0.0278 & 0.7340 & 0.0506 & 0.0285 & 0.7500 \\
\texttt{impresso}-en (10)    & 0.0372 & 0.0350 & 0.4000 & 0.0350 & 0.4000 & 0.0636 & 0.0335 & 0.5000 \\
\texttt{impresso}-fr (10)    & 0.0150 & 0.0113 & 0.6000 & 0.0102 & 0.8000 & 0.0527 & 0.0154 & 0.0000 \\
\texttt{impresso}-de (10)    & 0.0226 & 0.0209 & 0.4000 & 0.0169 & 0.9000 & 0.0340 & 0.0219 & 0.0000 \\
\texttt{dta19-l0} (100)      & 0.0045 & 0.0078 & $-$0.5300 & 0.0045 & 0.2000 & 0.0182 & 0.0049 & 0.0000 \\
\texttt{dta19-l1} (100)      & 0.0230 & 0.0231 & 0.1400 & 0.0197 & 0.4300 & 0.0775 & 0.0224 & 0.0600 \\
\texttt{dta19-l2} (100)      & 0.0570 & 0.0511 & 0.5000 & 0.0498 & 0.6200 & 0.1441 & 0.0567 & 0.1300 \\
\bottomrule
\end{tabular}
\end{table}
\fi

\paragraph{Per-span generation fails even with oracle spans.}
Given the exact error spans, \mbox{Qwen3-8B} generating one replacement per span is worse than no correction (cMER 0.0562 and preference $-0.6247$, against 0.0275 and 0) in every development cell. Of 13,167 oracle spans, 5,229 proposals were rejected (too long or without tags) and 4,677 echoed the OCR span; of the 3,261 that changed it, only 331 matched the reference, and 970 returned the OCR span followed by text from the right-hand context. A window interface removes this leakage but still scores below the unfiltered rewrite (0.0255 / 0.2627).

\paragraph{Vetoing a full-context rewrite helps.}
Applying the oracle span gate to the full zero-shot rewrite instead lowers cMER from 0.0247 to 0.0229 and raises the preference score from 0.4161 to 0.6501. The reverse span gate is worse than no correction (0.0292 / $-0.3181$), so the gain comes from which blocks are kept. Character granularity is slightly worse than blocks (0.0236 / 0.5681), and we use blocks. A probe with \mbox{Qwen3-32B} and thinking mode is reported in \apptabref{tab:probe}.

\iforacleappendix\else\begin{table}[t]
\centering
\caption{Development \emph{probe subset} (100 units, 90 in official cells; fixed per-cell allocation, seed 1234), \emph{oracle} analysis: weighted cMER / preference score. Not comparable to Table~\ref{tab:oracle}. --: not run.}
\label{tab:probe}
\small\setlength{\tabcolsep}{4pt}
\begin{tabular}{lcccc}
\toprule
 & \multicolumn{2}{c}{Qwen3-8B} & \multicolumn{2}{c}{Qwen3-32B} \\
\cmidrule(lr){2-3}\cmidrule(lr){4-5}
System & cMER & Pref & cMER & Pref \\
\midrule
No correction & 0.0265 & 0.0000 & 0.0265 & 0.0000 \\
Zero-shot rewrite (ZS) & 0.0238 & 0.5422 & 0.0284 & 0.6533 \\
ZS + oracle span gate (block) & 0.0222 & 0.7022 & 0.0178 & 0.8000 \\
ZS + oracle span gate (char.) & 0.0219 & 0.6222 & \textbf{0.0169} & \textbf{0.8133} \\
ZS + reverse span gate (block) & -- & -- & 0.0371 & $-$0.4244 \\
Oracle window rewrite & 0.0242 & 0.3289 & 0.0258 & 0.3733 \\
\quad thinking on, 1,024 tokens & 0.0250 & 0.3156 & -- & -- \\
\quad thinking on, 4,096 tokens & 0.0291 & $-$0.0889 & -- & -- \\
\bottomrule
\end{tabular}
\end{table}
\fi

\section{Learned Gate: Development Results}
\label{sec:results}

Development results for the two frozen gates on all five rewriters are in \apptabref{tab:dev}. No number here chose the gates or thresholds (Sect.~\ref{sec:gate}), but the rewriter list and the freeze were decided after these runs.

\paragraph{The gate helps every rewriter.}
The full gate lowers weighted cMER for all five rewriters, by 13\% (\mbox{Qwen3-8B}), 27\% (\mbox{Qwen3-32B}), 19\% (LoRA), 47\% (\mbox{Mistral-Small-24B}), and 38\% (\mbox{Qwen2.5-72B}), and it beats the oracle span gate on cMER in every row. In the paired bootstrap each of these differences is significant on both metrics ($p\le0.0008$), with $\Delta$cMER from $-0.0032$ $[-0.0044,-0.0021]$ for \mbox{Qwen3-8B} to $-0.0118$ $[-0.0170,-0.0075]$ for \mbox{Mistral-Small-24B}. It closes 68\%, 70\%, 50\%, 84\%, and 74\% of the cMER gap to the label oracle.

Ungated quality predicts gated quality poorly: \mbox{Mistral-Small-24B} over-edits (ungated 0.0251) but proposes twice as many helpful blocks as \mbox{Qwen3-8B} (12,240 against 6,084), and gated it is among the best (0.0133).

\paragraph{Transfer and log-probabilities.}
A gate trained on one Qwen3 rewriter's blocks and applied to the other loses at most 0.0004 cMER against the same-rewriter gate, and pooling both rewriters, chosen for its held-out ROC-AUC, is slightly worse on \mbox{Qwen3-8B} (0.0213 / 0.677 against 0.0211 / 0.700) and slightly better on \mbox{Qwen3-32B} (0.0160 against 0.0162). Removing every log-probability feature changes cMER by $-0.0002$ to $+0.0004$ on the two Qwen3 rewriters and lowers held-out ROC-AUC by 0.003--0.007. On \mbox{Mistral-Small-24B} and \mbox{Qwen2.5-72B}, the two rewriters unseen in training, the text-only gate is within 0.0005 cMER of the full gate but lower on preference by 0.02--0.06. The training-free log-probability thresholds are weak (held-out block ROC-AUC 0.62--0.63; on \mbox{Qwen3-8B} $\Delta$cMER $-0.0006$, $p=0.07$). Among the backbones, \gatebb and \mbox{Qwen3-1.7B}~\cite{qwen3} are similar and ahead of byT5-base~\cite{xue2022byt5} (held-out ROC-AUC 0.945, 0.949, and 0.928 on \mbox{Qwen3-8B} blocks), so we kept the smallest.

\section{Test Results}
\label{sec:test}

Results before Sect.~\ref{sec:phase2} come from look~1, the first frozen test run (Sect.~\ref{sec:setup}); label-oracle results read the test references and are not systems.

\begin{figure}[t]
\centering
\includegraphics{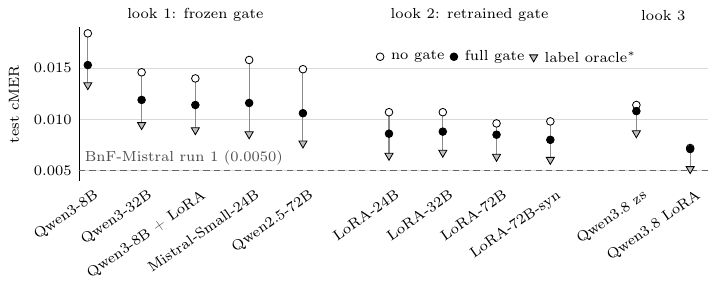}
\caption{Test cMER (official weighted mean, 590 units) of every single rewriter without gate, with the full gate, and with the label oracle ($^\ast$reads the reference; not a system). Look~1 uses the frozen gate, looks~2 and~3 the retrained one (Sect.~\ref{sec:phase2}); no correction is at 0.0226. Values in Table~\ref{tab:look2} and \protect\apptabref{tab:main}.}
\label{fig:gate}
\end{figure}

\paragraph{The frozen gates improve every rewriter.}
Both gates lower weighted cMER and raise the preference score for all five rewriters, and all 20 differences are significant at the resolution of the bootstrap ($p\approx0.0002$; Fig.~\ref{fig:gate} and \apptabref{tab:main}): the full gate lowers cMER by 0.0026--0.0044 and raises the preference score from 0.49--0.71 to 0.71--0.85. This includes \mbox{Mistral-Small-24B} and \mbox{Qwen2.5-72B}, whose outputs the gate never saw in training, and the LoRA rewriter; \mbox{Qwen2.5-72B} is from the same family as the training rewriters, so \mbox{Mistral-Small-24B} is the only one from another family. The text-only gate equals the full gate in cMER to four decimals for four rewriters and differs by 0.0001 for \mbox{Qwen3-32B}; its preference score is lower by 0.01--0.02 for the zero-shot rewriters. The gate thus needs only the input and output text, which, with the transfer results, is the sense in which it is rewriter-agnostic. Of the labeled gain that the label oracle keeps, the gates keep 79--82\%, and they achieve about 50--62\% of its cMER reduction. Block-level ROC-AUC on test is 0.869--0.908, lower than on held-out training blocks (0.949--0.952).

\paragraph{Ranking and comparison with the competition.}
The best look-1 system is \mbox{Qwen2.5-72B} with either gate (0.0106 / 0.8446 with the full gate). Before look~1 we had named two primary systems, \mbox{Mistral-Small-24B} with the full gate (best on development data) and \mbox{Qwen2.5-72B} with the text-only gate; they rank third and first. Against the published results~\cite{ehrmann2026hipe}, our best system is level with ungated BLOCR (0.0106 / 0.7028) on cMER, with a higher preference score, and behind all three BnF-Mistral runs (runs 1, 3, and 2: 0.0050, 0.0071, 0.0090). It is not state of the art.

\paragraph{Where the gate hurts.}
Per cell (details in \apptabref{tab:perset}), the full gate lowers cMER in 34 of the 40 rewriter--cell pairs. On \texttt{icdar2017}-fr it raises cMER for four of the five rewriters, by 0.0012--0.0023 with either gate, significantly for \mbox{Qwen3-32B} and \mbox{Mistral-Small-24B} ($p\le0.0026$). The label oracle roughly halves this cell's cMER for three rewriters (e.g.\ \mbox{Qwen3-32B} 0.0087 to 0.0050), so the gate rejects too many helpful edits on the one official cell without a development file. On \texttt{impresso}-en it raises cMER slightly for \mbox{Mistral-Small-24B} and \mbox{Qwen2.5-72B} (+0.0004 to +0.0010; significant only once, $p=0.02$).

\paragraph{Gating the competition systems.}
On the published test outputs (details in \apptabref{tab:participants}) the text-only gate raises the preference score of every run. It lowers cMER for BLOCR (0.0106 to 0.0096; lower in 6 of 8 cells, higher on \texttt{icdar2017}-fr and \texttt{impresso}-en), for both L3i runs, and for Zakaria-ENSIAS (0.0209 to 0.0121). It does not help BnF-Mistral on the primary metric: cMER rises for run~1 (0.0050 to 0.0065, higher in 7 of 8 cells) and run~3 (0.0071 to 0.0077) and is unchanged for run~2. The preference gain of run~1 comes mainly from \texttt{dta19-l0} (+0.33), where the gate vetoes edits on nearly clean pages. The global threshold was tuned on less accurate rewriters; when most of a system's edits are right, so are most vetoed ones.

\subsection{Second and Third Test Looks: Fine-Tuned Rewriters and Edit Pools}
\label{sec:phase2}
After look~1 we tried to close the gap to the winning run with stronger rewriters. This changes what is evaluated: the rewriters below are fine-tuned on the training split and their gated edits are pooled, so their scores mix the effect of the gate with those of fine-tuning and pooling. We report the gate's effect as gated against ungated output of the same rewriter. Selection is described in Sect.~\ref{sec:setup}.

\paragraph{Systems.} \emph{Rewriters:} LoRA adapters (rank 16, $\alpha=32$, two epochs, epoch chosen on held-out training units) on \mbox{Mistral-Small-24B}, \mbox{Qwen3-32B}, and \mbox{Qwen2.5-72B}, trained on 1,572 examples from the 70\% of training units set aside for LoRA training; a second \mbox{Qwen2.5-72B} adapter (LoRA-72B-syn) also sees 1,520 synthetic pairs made by corrupting public-domain and CC~BY-SA historical text and four \texttt{impresso-nzz} training units, with confusion statistics estimated from the LoRA training units. \emph{Gate:} the look-1 gate saw only Qwen3 blocks, so we retrained the same architecture, features, and rule on blocks from six rewriters: zero-shot \mbox{Qwen3-8B}, \mbox{Qwen3-32B}, \mbox{Mistral-Small-24B}, and \mbox{Qwen2.5-72B} on the whole training split, and the 24B and 32B LoRA rewriters on the 721 training units held out from LoRA training (one global threshold, \appref{app:look2}). This is \emph{not} the frozen look-1 gate: it has seen training-split outputs of every look-2 rewriter except the two 72B LoRA rewriters, and zero-shot outputs of their base model, so look~2 is not evidence for the rewriter-agnostic claim. \emph{Ensembles:} in ensemble~B, the primary system, the edit blocks of the four LoRA rewriters are pooled per unit and identical blocks merged; a block is a candidate if its full-gate score passes the threshold of its agreement level (0.779 if one rewriter proposes it, 0.310 for two, 0.012 for three or four); overlapping candidates are resolved by the higher score, and the accepted blocks are applied to the OCR text. Proposition~1(ii) no longer applies. The pool and the levels were selected by the official cMER of the actual ensemble on 360 held-out training units, the thresholds by labeled gain on the same units. Ensemble~A, built first, pools LoRA-24B, LoRA-32B, and zero-shot \mbox{Qwen2.5-72B}. On development data (reporting only) B reaches 0.0084 / 0.971, against 0.0089 / 0.971 for A ($p=0.12$); sampled candidates made every pool worse, and synthetic pairs gave no gain once gated (\appref{app:look2}).

\begin{table}[t]
\centering
\caption{Looks~2 and~3, test set (590 scored units): weighted cMER / preference score. Gates: the retrained gate (Sect.~\ref{sec:phase2}), unchanged in look~3. $\Delta$: full gate $-$ no gate, paired bootstrap, 95\% CI (look~2: $p\approx0.0002$ on cMER for all five, on preference $p\le0.011$; look~3: see text). $^\dagger$The gate was trained on this rewriter's training-split outputs. LoRA-24B/32B/72B: adapters on \mbox{Mistral-Small-24B}, \mbox{Qwen3-32B}, \mbox{Qwen2.5-72B}; ensembles A--C: text; zs: zero-shot, rerun with the look-1 prompt and decoding (look~1: 0.0149 / 0.6074).}
\label{tab:look2}
\scriptsize\setlength{\tabcolsep}{2.0pt}
\begin{tabular}{lcccc}
\toprule
Rewriter & No gate & Full gate & Text-only & $\Delta$cMER (full) \\
\midrule
\mbox{Qwen2.5-72B} zs$^\dagger$ & 0.0148 / 0.6107 & 0.0105 / 0.8343 & 0.0106 / 0.8324 & $-$0.0043 [$-$0.0055, $-$0.0034] \\
LoRA-24B$^\dagger$ & 0.0107 / 0.7980 & 0.0086 / 0.9169 & 0.0087 / 0.9100 & $-$0.0021 [$-$0.0026, $-$0.0016] \\
LoRA-32B$^\dagger$ & 0.0107 / 0.7961 & 0.0088 / 0.8980 & 0.0089 / 0.8913 & $-$0.0019 [$-$0.0023, $-$0.0015] \\
LoRA-72B & 0.0096 / 0.8409 & 0.0085 / 0.8944 & 0.0086 / 0.8974 & $-$0.0011 [$-$0.0016, $-$0.0007] \\
LoRA-72B-syn & 0.0098 / 0.9004 & 0.0080 / 0.9320 & 0.0081 / 0.9302 & $-$0.0018 [$-$0.0042, $-$0.0004] \\
\midrule
Ensemble A & -- & 0.0071 / 0.9244 & -- & -- \\
\textbf{Ensemble B} & -- & \textbf{0.0069 / 0.9282} & -- & -- \\
\midrule
\multicolumn{5}{l}{\emph{Look~3 (secondary, post hoc)}} \\
Qwen3.8 zs & 0.0114 / 0.7704 & 0.0108 / 0.8289 & 0.0108 / 0.8374 & $-$0.0006 [$-$0.0011, $-$0.0002] \\
Qwen3.8 LoRA & 0.0072 / 0.8928 & 0.0071 / 0.9250 & 0.0071 / 0.9372 & $-$0.0001 [$-$0.0003, +0.0002] \\
Ensemble C & -- & 0.0066 / 0.9296 & -- & -- \\
\bottomrule
\end{tabular}
\end{table}

\paragraph{The gate's effect.} With the rewriter held fixed, the retrained gate lowers cMER and raises the preference score for all five rewriters (Table~\ref{tab:look2}). The cMER reduction is 0.0011--0.0021 for the fine-tuned rewriters, against 0.0043 for zero-shot \mbox{Qwen2.5-72B}: a rewriter that already makes fewer harmful edits leaves the gate less to remove. It is smallest for LoRA-72B, one of the two rewriters new to the gate. The text-only gate is within 0.0001 of the full gate on cMER. Per cell, the gate lowers cMER in 32 of the 40 rewriter--cell pairs; all eight increases are on \texttt{icdar2017}-fr and \texttt{impresso}-en, the cells where the look-1 gate also hurt.

\paragraph{The effect of fine-tuning and pooling.} Most of the step from the best look-1 system (0.0106) to the ensemble (0.0069) does not come from the gate. Fine-tuning lowers the ungated cMER of \mbox{Qwen2.5-72B} from 0.0148 to 0.0096, and pooling lowers the cMER of the gated LoRA-72B, the pre-declared single-rewriter reference (best on development data, tied with LoRA-72B-syn, $p=0.94$), from 0.0085 to 0.0069 ($-0.0016$ [$-0.0021$, $-0.0012$], $p\approx0.0002$). On test the gated LoRA-72B-syn is the best single look-2 system (0.0080; $-0.0005$ against LoRA-72B, $p=0.020$); it was not bootstrapped against the ensemble. Because the agreement thresholds act on gate scores, the gate's share inside the ensemble is not separated. Ensemble B is slightly better than A (Table~\ref{tab:look2}; $-0.0003$ [$-0.0005$, $-0.0001$], $p=0.003$).

\paragraph{Comparison with the competition.} Scored with the official scorer against the published runs, the (unsubmitted) ensemble is, on cMER, behind BnF-Mistral run~1 ($+0.0019$ [+0.0014, +0.0024], $p\approx0.0002$), statistically tied with BnF-Mistral run~3, the second-ranked run ($-0.0002$ [$-0.0008$, +0.0004], $p=0.49$), and better than BnF-Mistral run~2 ($-0.0022$, $p\approx0.0002$), BLOCR ($-0.0038$, $p\approx0.0002$), and L3i. Its preference score, 0.928, is above that of every published run (not significantly for run~1, $p=0.08$). It is not state of the art, but among published runs only BnF-Mistral runs~1 and~3 are better or tied, and they build on continued pre-training on about 80B tokens of pre-1900 BnF documents~\cite{ehrmann2026hipe}, for which we found no public release; ours uses only public models and data. We therefore see the gate as complementary to domain pre-training, not a substitute (Sect.~\ref{sec:conclusion}). These comparisons rest on a second test look (Sect.~\ref{sec:setup}).

\paragraph{Third look (secondary).} A pre-declared third look tested \mbox{Qwen3.8-27B}, zero-shot and LoRA-tuned, under the unchanged look-2 gate, frozen a day before the first Qwen3.8 output; the model is new to the gate, but its family is not. On the zero-shot rewriter the gate lowers cMER ($p=0.002$) and raises the preference score ($+0.059$, $p=0.0006$); on the LoRA rewriter cMER is flat ($p=0.49$) and only the preference score rises ($+0.032$, $p=0.003$), while the label oracle (0.0087 and 0.0052) shows headroom the gate does not reach. Ensemble~C pools both Qwen3.8 rewriters with the two 72B adapters (two agreement levels) and was run only because it beat B on the held-out units (0.0073 against 0.0076), a rule fixed in advance; it is slightly better than B ($-0.0003$ [$-0.0005$, $-0.0000$], $p=0.018$) and behind BnF-Mistral run~1 ($+0.0017$), but it was designed after two looks; B remains our primary system (\appref{app:look2}).

\section{Conclusion}
\label{sec:conclusion}
For the rewriter and prompt studied, an edit gate is useful for vetoing the collateral edits of a full-context rewrite, not for driving generation span by span, and a small learned gate can do this vetoing. Trained on blocks from two Qwen3 rewriters and evaluated frozen, it improves all five rewriters we tested on both metrics, including two it never saw in training, also in a text-only variant; it helps three of the four competition teams but raises the cMER of the lowest-cMER system. In a second, disclosed test look, a retrained gate still improves LoRA-tuned rewriters, by less than it improves zero-shot ones, and pooling their gated edits gives a cMER of 0.0069, tied with the second-ranked competition run and behind the winner (0.0050). Most of that gain comes from fine-tuning and pooling. In a third, secondary look the same gate, unchanged, improves a newer zero-shot rewriter but only the preference score of its LoRA-tuned version. The gate is a safeguard for the rewriters we tested, not for the lowest-cMER published system, and not by itself a route to the state of the art; we see it as complementary to domain pre-training.

\subsection*{Future Work}
A single global threshold suits rewriters of moderate precision but raises the cMER of the two lowest-cMER systems we gated, BnF-Mistral runs~1 and~3 (+0.0015 and +0.0006); a gate that adapts to the rewriter's precision, through per-system or calibrated thresholds, could avoid this. The label oracle leaves headroom even for the strongest single rewriter: gated Qwen3.8 LoRA reaches 0.0071 against 0.0052 for the oracle (look~3, secondary), and since the oracle selects from the same blocks, closing this gap is a matter of better block scores. Pairing the gate with rewriters that use continued pre-training on historical text, as the winning system does, would test this complementarity, which our fixed-threshold gate did not show on BnF-Mistral's outputs. Finally, other OCR benchmarks and languages are needed to test the gate beyond \hipe.

\subsection*{Limitations}
Our evaluation is restricted to the text-only setting and to the three languages and the normalization of \hipe, whose information-retrieval-oriented scoring ignores case and punctuation and is not a measure of diplomatic fidelity~\cite{ehrmann2026hipe}. A veto can only remove edits, so the gate inherits the rewriter's recall. The global threshold costs cMER on \texttt{icdar2017}-fr, on \texttt{impresso}-en, and on the lowest-cMER competition system (Sect.~\ref{sec:test}). Looks~2 and~3 may be optimistic through adaptive overfitting (Sect.~\ref{sec:setup}). The look-2 gate is retrained on outputs of most of the rewriters it is applied to, and the ensemble needs the rewriters' log-probabilities, so neither supports the rewriter-agnostic claim, which rests on looks~1 and~3; the new rewriter of look~3 is from a model family the gate has seen. All results come from one rewriter run and one gate seed per configuration and look, and small cells (30 test units per \texttt{dta19} level) give coarse per-cell results. The bound of Eq.~\eqref{eq:bound} limits damage but does not guarantee improvement.

\ifanonymous\else
\subsection*{Code and Data Availability}
\label{sec:availability}
Code, gate configurations, frozen-configuration checksums, and run logs are released at \REPOURL\ under the Apache-2.0 license. The trained artifacts, namely the four gates (the frozen look-1 and retrained look-2 gates, each full and text-only, based on \gatebb) and the LoRA adapters \mbox{Qwen3-8B} + LoRA (look~1), \mbox{LoRA-24B} (\mbox{Mistral-Small-24B}), \mbox{LoRA-32B} (\mbox{Qwen3-32B}), \mbox{LoRA-72B} (\mbox{Qwen2.5-72B}), and Qwen3.8 LoRA (\mbox{Qwen3.8-27B}), are released separately on Hugging Face (\HFURL) under CC~BY-NC-SA~4.0, for non-commercial research use, because of the licenses of the training data; the base models keep their own licenses, and \mbox{LoRA-72B} is also subject to the Qwen license agreement. The \mbox{LoRA-72B-syn} adapter and the synthetic training pairs are withheld, because the clean texts they are built from include share-alike and non-commercial sources that we do not redistribute (German Wikisource transcriptions under CC~BY-SA~3.0 and \texttt{impresso-nzz} ground truth under CC~BY-NC~4.0); the repository README gives the recipe to regenerate them. The \hipe data, the official scorer, and the published competition outputs are available from the organizers~\cite{ehrmann2026hipe}; we do not redistribute them.
\fi

\ifanonymous
\begin{credits}
\subsubsection{Declaration of AI-assisted writing and tools.} During the preparation of this work, the author(s) used AI assistants, Grok Bot (SpaceXAI) and Claude (Anthropic), to help with language refinement, manuscript organization, drafting portions of the text from author-provided research materials, and implementing and running experiment code under the author(s)' direction. The author(s) designed the study, reviewed and verified all content, code and results, and take(s) full responsibility for the scientific accuracy, originality and conclusions of this manuscript.
\end{credits}
\else
\begin{credits}
\subsubsection{\ackname} The author thanks the Chambers-Grundy Center for Transformative Neuroscience and the Department of Brain Health, Kirk Kerkorian School of Medicine, University of Nevada, Las Vegas, for providing the GPU computing resources used in this work.

\subsubsection{\discintname} The author has no competing interests to declare.

\subsubsection{Declaration of AI-assisted writing and tools.} During the preparation of this work, the author used AI assistants, Grok Bot (SpaceXAI) and Claude (Anthropic), to help with language refinement, manuscript organization, drafting portions of the text from author-provided research materials, and implementing and running experiment code under the author's direction. The author designed the study, reviewed and verified all content, code and results, and takes full responsibility for the scientific accuracy, originality and conclusions of this manuscript.
\end{credits}
\fi

\bibliographystyle{splncs04}
\bibliography{refs}

\ifextended
\appendix
\section{Proof and Checks for Proposition~1}
\label{app:proof}
\noindent\emph{Proof.} (i) holds by construction. (ii) The operations of $\Omega$ inside block $i$ form a minimum-cost script from $B_i$ to $B'_i$; otherwise replacing them would shorten $\Omega$. Hence $d(x,z)=\sum_i d(B_i,B'_i)$. Applying the accepted blocks to $x$ and the vetoed ones to $\hat{y}$ gives $d(x,\hat{y})\le\sum_{i\in A} d(B_i,B'_i)$ and $d(\hat{y},z)\le\sum_{i\notin A} d(B_i,B'_i)$. The two right-hand sides add up to $d(x,z)\le d(x,\hat{y})+d(\hat{y},z)$, so both inequalities are equalities. (iii) follows from the triangle inequality for $d$ on normalized strings. $\square$

The same argument applies when the gate accepts single operations of $\Omega$ rather than whole blocks. Statement (ii) concerns raw strings; its normalized analogue does not hold in general, because $\norm$ does not act character by character: inserting one letter between \textit{¬} and the following line break prevents the word parts from being joined, so one raw insertion becomes three normalized insertions. Part (iii) holds for any output; its use here is that the veto controls the distance terms, since by (ii) the output moves away from $x$ only through accepted blocks and away from $z$ only through vetoed ones. In practice $\Omega$ is the script returned by \texttt{rapidfuzz} \texttt{Levenshtein.editops}. On all 4,488 logged development and probe outputs its length equals the Levenshtein distance computed by an independent implementation (which shows minimality, not uniqueness), and (i)--(iii) hold for every tested block subset (22 per output); the normalized analogue of (ii) fails for 1,375 of the 98,736 subsets. Eq.~\eqref{eq:bound} bounds error counts, not $\cmer$ itself, whose denominator $|\norm(y^{\star})|+I$ depends on the aligner's tie-breaking; for a non-empty reference it implies only $\cmer(\hat{y}) \le E(\hat{y})/|\norm(y^{\star})|$.

\section{Additional Development-Set Tables}
\label{app:oracle}
\iforacleappendix
Tables~\ref{tab:oracle}--\ref{tab:probe} supplement Sect.~\ref{sec:oracle}, and Table~\ref{tab:dev} supplements Sect.~\ref{sec:results}.
\begin{table}[t]
\centering
\caption{Development set, \emph{oracle} analysis (earlier \mbox{Qwen3-8B} zero-shot run; see Sect.~\ref{sec:systems}): weighted means over the seven official development cells (518 units). Oracle systems read error locations from the reference and are not valid systems.}
\label{tab:oracle}
\small\setlength{\tabcolsep}{5pt}
\begin{tabular}{lcc}
\toprule
System & cMER $\downarrow$ & Pref $\uparrow$ \\
\midrule
No correction & 0.0275 & 0.0000 \\
Zero-shot rewrite (ZS) & 0.0247 & 0.4161 \\
\midrule
Oracle per-span generation (earlier design) & 0.0562 & $-$0.6247 \\
Oracle window rewrite & 0.0255 & 0.2627 \\
\midrule
ZS + oracle span gate (block) & \textbf{0.0229} & \textbf{0.6501} \\
ZS + oracle span gate (character) & 0.0236 & 0.5681 \\
ZS + reverse span gate (block; sanity check) & 0.0292 & $-$0.3181 \\
\bottomrule
\end{tabular}
\end{table}

\begin{table}[t]
\centering
\caption{Development set, \emph{oracle} analysis: per-cell micro cMER and preference score (earlier \mbox{Qwen3-8B} zero-shot run). ZS+OSG: zero-shot rewrite with the oracle span gate (block). OPG: oracle per-span generation. OW: oracle window rewrite. Units per cell in parentheses.}
\label{tab:oraclecell}
\scriptsize\setlength{\tabcolsep}{2.6pt}
\begin{tabular}{lcccccccc}
\toprule
 & No corr. & \multicolumn{2}{c}{ZS} & \multicolumn{2}{c}{ZS + OSG} & OPG & \multicolumn{2}{c}{OW} \\
\cmidrule(lr){3-4}\cmidrule(lr){5-6}\cmidrule(lr){8-9}
Cell & cMER & cMER & Pref & cMER & Pref & cMER & cMER & Pref \\
\midrule
\texttt{icdar2017}-en (188)  & 0.0343 & 0.0288 & 0.6436 & 0.0278 & 0.7340 & 0.0506 & 0.0285 & 0.7500 \\
\texttt{impresso}-en (10)    & 0.0372 & 0.0350 & 0.4000 & 0.0350 & 0.4000 & 0.0636 & 0.0335 & 0.5000 \\
\texttt{impresso}-fr (10)    & 0.0150 & 0.0113 & 0.6000 & 0.0102 & 0.8000 & 0.0527 & 0.0154 & 0.0000 \\
\texttt{impresso}-de (10)    & 0.0226 & 0.0209 & 0.4000 & 0.0169 & 0.9000 & 0.0340 & 0.0219 & 0.0000 \\
\texttt{dta19-l0} (100)      & 0.0045 & 0.0078 & $-$0.5300 & 0.0045 & 0.2000 & 0.0182 & 0.0049 & 0.0000 \\
\texttt{dta19-l1} (100)      & 0.0230 & 0.0231 & 0.1400 & 0.0197 & 0.4300 & 0.0775 & 0.0224 & 0.0600 \\
\texttt{dta19-l2} (100)      & 0.0570 & 0.0511 & 0.5000 & 0.0498 & 0.6200 & 0.1441 & 0.0567 & 0.1300 \\
\bottomrule
\end{tabular}
\end{table}

\begin{table}[t]
\centering
\caption{Development \emph{probe subset} (100 units, 90 in official cells; fixed per-cell allocation, seed 1234), \emph{oracle} analysis: weighted cMER / preference score. Not comparable to Table~\ref{tab:oracle}. --: not run.}
\label{tab:probe}
\small\setlength{\tabcolsep}{4pt}
\begin{tabular}{lcccc}
\toprule
 & \multicolumn{2}{c}{Qwen3-8B} & \multicolumn{2}{c}{Qwen3-32B} \\
\cmidrule(lr){2-3}\cmidrule(lr){4-5}
System & cMER & Pref & cMER & Pref \\
\midrule
No correction & 0.0265 & 0.0000 & 0.0265 & 0.0000 \\
Zero-shot rewrite (ZS) & 0.0238 & 0.5422 & 0.0284 & 0.6533 \\
ZS + oracle span gate (block) & 0.0222 & 0.7022 & 0.0178 & 0.8000 \\
ZS + oracle span gate (char.) & 0.0219 & 0.6222 & \textbf{0.0169} & \textbf{0.8133} \\
ZS + reverse span gate (block) & -- & -- & 0.0371 & $-$0.4244 \\
Oracle window rewrite & 0.0242 & 0.3289 & 0.0258 & 0.3733 \\
\quad thinking on, 1,024 tokens & 0.0250 & 0.3156 & -- & -- \\
\quad thinking on, 4,096 tokens & 0.0291 & $-$0.0889 & -- & -- \\
\bottomrule
\end{tabular}
\end{table}

\begin{table}[htbp]
\centering
\caption{Development set: weighted cMER / preference score over the seven official cells (518 units). Full and text-only: the frozen pooled \gatebb gates. Columns marked $^\ast$ read the reference and are not valid systems. No correction: 0.0275 / 0.000. n/r: not run.}
\label{tab:dev}
\scriptsize\setlength{\tabcolsep}{2.4pt}
\begin{tabular}{lccccc}
\toprule
Rewriter & No gate & Full gate & Text-only gate & Span oracle$^\ast$ & Label oracle$^\ast$ \\
\midrule
\mbox{Qwen3-8B}         & 0.0245 / 0.461 & 0.0213 / 0.677 & 0.0211 / 0.679 & 0.0228 / 0.656 & 0.0198 / 0.745 \\
\mbox{Qwen3-32B}        & 0.0220 / 0.600 & 0.0160 / 0.810 & 0.0161 / 0.811 & 0.0186 / 0.793 & 0.0135 / 0.859 \\
\mbox{Qwen3-8B} + LoRA  & 0.0176 / 0.677 & 0.0143 / 0.898 & n/r            & 0.0159 / 0.853 & 0.0108 / 0.970 \\
\mbox{Mistral-Small-24B} & 0.0251 / 0.486 & 0.0133 / 0.896 & 0.0138 / 0.837 & 0.0195 / 0.732 & 0.0110 / 0.944 \\
\mbox{Qwen2.5-72B}      & 0.0216 / 0.563 & 0.0134 / 0.833 & 0.0133 / 0.815 & 0.0162 / 0.802 & 0.0105 / 0.910 \\
\bottomrule
\end{tabular}
\end{table}

\fi
\section{Additional Test-Set Tables}
\label{app:testpref}
\begin{table}[htbp]
\centering
\caption{Test set, our rewriters: weighted micro cMER and preference score over the eight official cells (590 scored units). Full and text-only: the frozen gates. LO$^\ast$: label oracle (reads the reference). Lower part: paired-bootstrap difference to the ungated rewriter (95\% CIs for the text-only gate in Table~\ref{tab:textonlyci}); $p\approx0.0002$, the smallest attainable value, for all 20 differences. No correction: 0.0226 / 0.0000.}
\label{tab:main}
\scriptsize\setlength{\tabcolsep}{3pt}
\begin{tabular}{lcccccccc}
\toprule
 & \multicolumn{2}{c}{No gate} & \multicolumn{2}{c}{Full gate} & \multicolumn{2}{c}{Text-only gate} & \multicolumn{2}{c}{LO$^\ast$} \\
Rewriter & cMER$\downarrow$ & Pref$\uparrow$ & cMER & Pref & cMER & Pref & cMER & Pref \\
\midrule
\mbox{Qwen3-8B}          & 0.0184 & 0.4946 & 0.0153 & 0.7124 & 0.0153 & 0.6950 & 0.0134 & 0.7780 \\
\mbox{Qwen3-32B}         & 0.0146 & 0.6433 & 0.0119 & 0.8228 & 0.0120 & 0.8083 & 0.0095 & 0.8844 \\
\mbox{Qwen3-8B} + LoRA   & 0.0140 & 0.7061 & 0.0114 & 0.8541 & 0.0114 & 0.8535 & 0.0090 & 0.9274 \\
\mbox{Mistral-Small-24B} & 0.0158 & 0.6002 & 0.0116 & 0.8461 & 0.0116 & 0.8289 & 0.0086 & 0.9230 \\
\mbox{Qwen2.5-72B}       & 0.0149 & 0.6074 & \textbf{0.0106} & 0.8446 & \textbf{0.0106} & 0.8281 & 0.0077 & 0.9132 \\
\bottomrule
\end{tabular}\\[3pt]
\begin{tabular}{lcccc}
\toprule
 & \multicolumn{2}{c}{Full gate $-$ no gate} & \multicolumn{2}{c}{Text-only $-$ no gate} \\
Rewriter & $\Delta$cMER [95\% CI] & $\Delta$Pref [95\% CI] & $\Delta$cMER & $\Delta$Pref \\
\midrule
\mbox{Qwen3-8B}          & $-$0.0031 [$-$0.0043, $-$0.0021] & +0.218 [0.171, 0.267] & $-$0.0030 & +0.200 \\
\mbox{Qwen3-32B}         & $-$0.0027 [$-$0.0032, $-$0.0022] & +0.179 [0.138, 0.222] & $-$0.0026 & +0.165 \\
\mbox{Qwen3-8B} + LoRA   & $-$0.0026 [$-$0.0036, $-$0.0018] & +0.148 [0.105, 0.193] & $-$0.0027 & +0.147 \\
\mbox{Mistral-Small-24B} & $-$0.0041 [$-$0.0058, $-$0.0028] & +0.246 [0.193, 0.302] & $-$0.0042 & +0.229 \\
\mbox{Qwen2.5-72B}       & $-$0.0044 [$-$0.0057, $-$0.0034] & +0.237 [0.188, 0.289] & $-$0.0043 & +0.221 \\
\bottomrule
\end{tabular}
\end{table}

\begin{table}[htbp]
\centering
\caption{Test set, per cell: micro cMER $\times100$ of the OCR (Base) and the best competition result per cell (Best; BnF-Mistral run~1 unless noted), both copied from Table~7 of~\cite{ehrmann2026hipe}, and of each rewriter without $\rightarrow$ with the frozen full gate (LoRA: \mbox{Qwen3-8B} + LoRA). Units: scored units. Weighted Best: BnF-Mistral run~1 overall. The \texttt{dta19} cells have weight $1/3$. Shaded: the gate raises cMER.}
\label{tab:perset}
\scriptsize\setlength{\tabcolsep}{2.2pt}
\newcommand{\h}[1]{\colorbox{black!12}{#1}}
\begin{tabular}{lrccccccc}
\toprule
Cell & Units & Base & Best & \mbox{Qwen3-8B} & \mbox{Qwen3-32B} & LoRA & Mistral & Qwen2.5 \\
\midrule
\texttt{dta19-l0}       & 30  & 0.40 & 0.29$^a$ & 0.88$\to$0.40 & 1.19$\to$0.34 & 0.45$\to$0.28 & 1.48$\to$0.34 & 1.27$\to$0.33 \\
\texttt{dta19-l1}       & 30  & 2.40 & 0.54 & 3.21$\to$1.55 & 1.87$\to$1.25 & 1.36$\to$1.08 & 1.58$\to$1.17 & 2.08$\to$0.95 \\
\texttt{dta19-l2}       & 30  & 5.46 & 0.82 & 5.03$\to$4.06 & 4.25$\to$2.94 & 3.43$\to$2.53 & 5.51$\to$2.92 & 4.33$\to$2.35 \\
\texttt{icdar2017}-en   & 100 & 2.60 & 0.44 & 1.71$\to$1.59 & 1.22$\to$0.96 & 1.26$\to$1.09 & 1.58$\to$0.93 & 1.65$\to$0.79 \\
\texttt{icdar2017}-fr   & 100 & 1.84 & 0.40 & \h{1.23$\to$1.36} & \h{0.87$\to$1.10} & 1.33$\to$1.29 & \h{0.89$\to$1.11} & \h{0.91$\to$1.05} \\
\texttt{impresso}-de    & 100 & 2.96 & 0.58 & 2.47$\to$1.87 & 2.05$\to$1.65 & 1.94$\to$1.47 & 2.11$\to$1.65 & 1.93$\to$1.55 \\
\texttt{impresso}-en    & 100 & 1.70 & 0.49$^b$ & 1.20$\to$1.14 & 1.04$\to$0.93 & 0.86$\to$0.84 & \h{0.83$\to$0.87} & \h{0.74$\to$0.82} \\
\texttt{impresso}-fr    & 100 & 1.69 & 0.44 & 1.36$\to$1.19 & 1.17$\to$1.01 & 1.28$\to$0.87 & 1.20$\to$0.95 & 1.18$\to$0.92 \\
\midrule
Weighted                & 590 & 2.26 & 0.50 & 1.84$\to$1.53 & 1.46$\to$1.19 & 1.40$\to$1.14 & 1.58$\to$1.16 & 1.49$\to$1.06 \\
\bottomrule
\end{tabular}\\[2pt]
{\footnotesize $^a$ BnF-Mistral run~2. $^b$ BLOCR run~1 is statistically tied~\cite{ehrmann2026hipe}.}
\end{table}

\begin{table}[htbp]
\centering
\caption{Test set, competition systems: weighted cMER and preference score of the published run~\cite{ehrmann2026hipe} (reproduced with the official scorer) $\to$ the same outputs after the frozen text-only gate, and the paired-bootstrap cMER difference with 95\% CI and $p$ for both metrics. Means over the cells each run covers (in parentheses). The gate leaves the no-correction baseline unchanged.}
\label{tab:participants}
\scriptsize\setlength{\tabcolsep}{1.3pt}
\begin{tabular}{lcccc}
\toprule
Run (cells) & cMER & Pref & $\Delta$cMER [95\% CI] & $p$ (cMER / Pref) \\
\midrule
BnF-Mistral 1 (8) & 0.0050$\to$0.0065 & 0.9000$\to$0.9252 & +0.0015 [+0.0012, +0.0019] & 0.0002 / 0.0090 \\
BnF-Mistral 2 (8) & 0.0090$\to$0.0090 & 0.8739$\to$0.9070 & $-$0.0001 [$-$0.0004, +0.0003] & 0.69 / 0.0054 \\
BnF-Mistral 3 (8) & 0.0071$\to$0.0077 & 0.8737$\to$0.9196 & +0.0006 [+0.0003, +0.0008] & 0.0002 / 0.0002 \\
BLOCR 1 (8) & 0.0106$\to$0.0096 & 0.7028$\to$0.8596 & $-$0.0011 [$-$0.0016, $-$0.0006] & 0.0002 / 0.0002 \\
L3i 1 (8) & 0.0176$\to$0.0157 & 0.3591$\to$0.4902 & $-$0.0018 [$-$0.0027, $-$0.0012] & 0.0002 / 0.0002 \\
L3i 2 (3) & 0.0221$\to$0.0199 & $-$0.0933$\to$0.0167 & $-$0.0022 [$-$0.0028, $-$0.0017] & 0.0002 / 0.0002 \\
Zakaria-ENS.\ 1 (7) & 0.0209$\to$0.0121 & 0.4176$\to$0.7631 & $-$0.0088 [$-$0.0121, $-$0.0061] & 0.0002 / 0.0002 \\
\bottomrule
\end{tabular}
\end{table}

\begin{table}[htbp]
\centering
\caption{Test set, per cell: preference score of each rewriter without $\rightarrow$ with the frozen full gate (590 scored units).}
\label{tab:testpref}
\scriptsize\setlength{\tabcolsep}{2.6pt}
\begin{tabular}{lrccccc}
\toprule
Cell & Units & \mbox{Qwen3-8B} & \mbox{Qwen3-32B} & LoRA-8B & Mistral-24B & \mbox{Qwen2.5-72B} \\
\midrule
\texttt{dta19-l0}     & 30  & $-$0.70$\to$0.00 & $-$0.97$\to$0.30 & 0.30$\to$0.53 & $-$0.47$\to$0.30 & $-$0.63$\to$0.33 \\
\texttt{dta19-l1}     & 30  & 0.20$\to$0.97 & 0.73$\to$1.00 & 0.83$\to$1.00 & 0.73$\to$1.00 & 0.47$\to$1.00 \\
\texttt{dta19-l2}     & 30  & 0.43$\to$0.97 & 0.53$\to$1.00 & 0.87$\to$1.00 & 0.67$\to$1.00 & 0.60$\to$1.00 \\
\texttt{icdar2017}-en & 100 & 0.70$\to$0.76 & 0.87$\to$0.94 & 0.91$\to$0.93 & 0.70$\to$0.94 & 0.78$\to$0.98 \\
\texttt{icdar2017}-fr & 100 & 0.85$\to$0.84 & 0.96$\to$0.94 & 0.78$\to$0.84 & 0.94$\to$0.98 & 0.92$\to$0.96 \\
\texttt{impresso}-de  & 100 & 0.42$\to$0.80 & 0.68$\to$0.80 & 0.52$\to$0.87 & 0.57$\to$0.94 & 0.61$\to$0.81 \\
\texttt{impresso}-en  & 100 & 0.60$\to$0.67 & 0.70$\to$0.83 & 0.71$\to$0.81 & 0.68$\to$0.77 & 0.83$\to$0.88 \\
\texttt{impresso}-fr  & 100 & 0.42$\to$0.56 & 0.55$\to$0.66 & 0.65$\to$0.83 & 0.40$\to$0.68 & 0.36$\to$0.66 \\
\bottomrule
\end{tabular}
\end{table}

\begin{table}[htbp]
\centering
\caption{Test set: paired-bootstrap difference between the text-only gate and no gate (95\% CI; $p\approx0.0002$ for all).}
\label{tab:textonlyci}
\scriptsize\setlength{\tabcolsep}{4pt}
\begin{tabular}{lcc}
\toprule
Rewriter & $\Delta$cMER [95\% CI] & $\Delta$Pref [95\% CI] \\
\midrule
\mbox{Qwen3-8B}          & $-$0.0030 [$-$0.0042, $-$0.0021] & +0.200 [0.154, 0.248] \\
\mbox{Qwen3-32B}         & $-$0.0026 [$-$0.0031, $-$0.0022] & +0.165 [0.124, 0.207] \\
\mbox{Qwen3-8B} + LoRA   & $-$0.0027 [$-$0.0037, $-$0.0019] & +0.147 [0.106, 0.192] \\
\mbox{Mistral-Small-24B} & $-$0.0042 [$-$0.0058, $-$0.0029] & +0.229 [0.177, 0.284] \\
\mbox{Qwen2.5-72B}       & $-$0.0043 [$-$0.0056, $-$0.0033] & +0.221 [0.172, 0.272] \\
\bottomrule
\end{tabular}
\end{table}

\section{Reproducibility and Harness Changes}
\label{app:repro}
\paragraph{Frozen artifacts.} Both gates use the \gatebb encoder at a fixed Hugging Face snapshot, bf16 inference on the GPU, and the thresholds $\tau=0.5064541101455688$ (full) and $0.4989624172449112$ (text-only). The SHA-256 checksums of both gate checkpoints and of the LoRA adapter were recomputed and compared with the frozen configuration immediately before the test run; all matched (the adapter's checksum had not been recorded at freezing time and was recorded then for the first time). Gate inference on the CPU in fp32 flips a few blocks near the threshold on development data, so the frozen run used the GPU path.

\paragraph{Order of the single run.} Checksum verification; the frozen script for the four zero-shot rewriters and the competition outputs, run once; the LoRA step, run once; the bootstrap; the report. No step was rerun, no run directory was deleted, and no threshold, gate, prompt, decoding parameter, or configuration file was changed.

\paragraph{Harness changes made before the affected step ran.}
(1)~The frozen script located each gated competition file for scoring with a file pattern that omitted the team name; with several teams per cell it could have scored another team's file. The development dry run had only one ``team'' and could not reveal this. The pattern was replaced by the exact file name the gating tool writes. (2)~The rewriter client sent the switch that disables Qwen3's thinking mode only to model names containing \emph{qwen3}, a rule added after the development LoRA run. The LoRA adapter is served under another name, so its test run would have had thinking on, unlike its development run. A flag forcing the switch was added and used only for the LoRA step; its completion-token count (269,886 against 268,291 for plain \mbox{Qwen3-8B}) and zero fallbacks are consistent with thinking off. (3)~The gating tool aborted on any difference between a submission's OCR text and the reference OCR. In the released files this happens for 36 submission records (typographic dashes, one extra space, and three truncated or duplicated records in one team's DTA files), all in units the scorer excludes. The tool now uses the submission's own OCR text as the block source for such records and counts them; scores are unaffected because these units are not scored.

\paragraph{Other bookkeeping.} The bootstrap tool was extended to drop scorer-excluded units, to refuse the test split without an explicit confirmation, and to align external submissions by document identifier; its output on development data is bit-identical to the earlier version. The evaluation repository contains three backup files next to the submissions; they are not submissions and were excluded, so 58 submission files were scored, as in the organizers' table.

\section{Second and Third Test Looks: Details}
\label{app:look2}

\paragraph{Gate.} The look-2 gate was fitted on 370,049 blocks with 91,340 held out; its global thresholds are 0.3757 (full) and 0.2967 (text-only). Thresholds by edit rate, language, or a system-confidence proxy were no better than one global threshold on cross-fitted held-out data. Counting the two 72B adapters, which share base model and real training data, as two rewriters is a modeling choice.

\paragraph{Development.} All numbers are weighted over the seven official development cells. Ungated, the 24B and 32B LoRA rewriters reach 0.0139 / 0.856 and 0.0132 / 0.853 (zero-shot: 0.0251 / 0.486 and 0.0220 / 0.600), level with the best gated look-1 system, \mbox{Mistral-Small-24B} with the look-1 gate (0.0133 / 0.896). The retrained gate lowers cMER for all nine rewriters it was checked on ($p\approx0.0002$); the frozen look-1 gate improves the 24B and 32B LoRA rewriters to 0.0114 and 0.0117, and the retrained gate is significantly better than it only for LoRA-32B (0.0110 against 0.0117, $p=0.0006$; slightly worse, not significantly, on zero-shot \mbox{Mistral-Small-24B}). The 72B LoRA rewriter reaches 0.0118 / 0.891 ungated and 0.0099 / 0.969 gated; the 72B-syn rewriter 0.0116 / 0.944 and 0.0099 / 0.964. Ensemble A (LoRA-24B, LoRA-32B, zero-shot \mbox{Qwen2.5-72B}; two agreement levels, thresholds 0.756 and 0.079) reaches 0.0089 / 0.971; the frozen ensemble B 0.0084 / 0.971 ($-0.0005$ [$-0.0011$, +0.0002], $p=0.12$; against gated \mbox{LoRA-72B} $-0.0015$, $p\approx0.0002$). \emph{Sampled candidates:} four rewrites at temperature 0.7 (seeds 1235--1238) from LoRA-72B and LoRA-32B were added to the pool. A labeled-gain criterion that ignores conflicts between overlapping blocks favored them, but the official cMER of the actual ensembles was worse with them in every pool, on held-out training units (0.0081--0.0085 against 0.0079) and on development data (0.0091--0.0098 against 0.0086); with all eight samples the ensemble reaches 0.0098 / 0.885 (against ensemble A $+0.0009$, $p=0.07$, and preference $-0.087$, $p=0.002$). From then on, pools and agreement levels were selected by the official cMER of the actual ensemble (thresholds still by labeled gain). This change was decided after the development scores of these candidate ensembles had been computed in the same batch; the held-out and development rankings agree. \emph{Synthetic pairs:} 1,520 pairs from public-domain Project Gutenberg texts, German Wikisource (\emph{Die Gartenlaube}, CC~BY-SA~3.0), and four \texttt{impresso-nzz} training units (CC~BY-NC~4.0~\cite{ehrmann2026hipe}), corrupted with per-family character confusion statistics from the LoRA training units, improve the ungated 72B rewriter slightly (0.0118 to 0.0116, $p=0.69$; preference +0.053, $p=0.03$) and not the gated one (0.0099 in both cases). The context-free confusion model is a likely limit.

\paragraph{Deviations.} The 72B adapters were trained in bf16 with the base model sharded over two GPUs instead of with 4-bit quantization (no library support on the server); the recipe was otherwise unchanged. Epoch selection on held-out training units picked epoch~2 for LoRA-24B although epoch~1 has the lower development cMER (0.0135 against 0.0139); this was not revisited.

\begin{table}[htbp]
\centering
\caption{Look~2, test set, per cell: micro cMER $\times100$ of the OCR and of each rewriter without $\rightarrow$ with the retrained full gate, and of the four-rewriter ensemble (Ens.). Q72: zero-shot \mbox{Qwen2.5-72B}. Shaded: the gate raises cMER.}
\label{tab:look2cell}
\newcommand{\h}[1]{\colorbox{black!12}{#1}}
\scriptsize\setlength{\tabcolsep}{2.2pt}
\begin{tabular}{lrccccccc}
\toprule
Cell & Units & OCR & LoRA-24B & LoRA-32B & LoRA-72B & 72B-syn & Q72 & Ens. \\
\midrule
\texttt{icdar2017}-en & 100 & 2.60 & 0.97$\to$0.81 & 0.97$\to$0.83 & 0.88$\to$0.79 & 0.83$\to$0.79 & 1.57$\to$0.82 & 0.61 \\
\texttt{icdar2017}-fr & 100 & 1.84 & 1.30$\to$1.17 & \h{1.12$\to$1.13} & \h{1.05$\to$1.09} & \h{0.97$\to$1.03} & \h{0.89$\to$1.00} & 0.94 \\
\texttt{impresso}-en & 100 & 1.70 & \h{0.69$\to$0.71} & 0.73$\to$0.70 & \h{0.66$\to$0.69} & \h{0.64$\to$0.72} & \h{0.74$\to$0.83} & 0.63 \\
\texttt{impresso}-fr & 100 & 1.69 & 0.70$\to$0.59 & 0.65$\to$0.64 & 0.78$\to$0.66 & 1.33$\to$0.61 & 1.19$\to$0.94 & 0.55 \\
\texttt{impresso}-de & 100 & 2.96 & 1.58$\to$1.09 & 1.58$\to$1.10 & 1.26$\to$0.98 & 1.08$\to$0.90 & 1.92$\to$1.50 & 0.77 \\
\texttt{dta19-l0} & 30 & 0.40 & 0.30$\to$0.21 & 0.47$\to$0.25 & 0.31$\to$0.20 & 0.30$\to$0.20 & 1.28$\to$0.33 & 0.21 \\
\texttt{dta19-l1} & 30 & 2.40 & 0.74$\to$0.54 & 0.95$\to$0.64 & 0.78$\to$0.62 & 0.82$\to$0.63 & 2.09$\to$0.95 & 0.49 \\
\texttt{dta19-l2} & 30 & 5.46 & 2.50$\to$1.59 & 2.77$\to$1.83 & 2.31$\to$1.82 & 1.91$\to$1.39 & 4.33$\to$2.30 & 1.17 \\
\midrule
Weighted & 590 & 2.26 & 1.07$\to$0.86 & 1.07$\to$0.88 & 0.96$\to$0.85 & 0.98$\to$0.80 & 1.48$\to$1.05 & 0.69 \\
\bottomrule
\end{tabular}
\end{table}

\begin{table}[htbp]
\centering
\caption{Look~2, test set: the four-rewriter ensemble minus each published run (ungated), paired bootstrap over the cells the run covers. Negative $\Delta$cMER and positive $\Delta$Pref favor the ensemble; preference CIs are omitted for space.}
\label{tab:look2part}
\scriptsize\setlength{\tabcolsep}{3pt}
\begin{tabular}{lccc}
\toprule
Run (cells) & cMER (run $\to$ ens.) & $\Delta$cMER [95\% CI], $p$ & $\Delta$Pref, $p$ \\
\midrule
BnF-Mistral 1 (8) & 0.0050 $\to$ 0.0069 & +0.0019 [+0.0014, +0.0024], 0.0002 & +0.028, 0.08 \\
BnF-Mistral 3 (8) & 0.0071 $\to$ 0.0069 & $-$0.0002 [$-$0.0008, +0.0004], 0.49 & +0.054, 0.0006 \\
BnF-Mistral 2 (8) & 0.0090 $\to$ 0.0069 & $-$0.0022 [$-$0.0027, $-$0.0016], 0.0002 & +0.054, 0.0016 \\
BLOCR 1 (8) & 0.0107 $\to$ 0.0069 & $-$0.0038 [$-$0.0046, $-$0.0031], 0.0002 & +0.225, 0.0002 \\
L3i 1 (8) & 0.0176 $\to$ 0.0069 & $-$0.0107 [$-$0.0122, $-$0.0094], 0.0002 & +0.569, 0.0002 \\
L3i 2 (3) & 0.0221 $\to$ 0.0065 & $-$0.0156 [$-$0.0172, $-$0.0141], 0.0002 & +1.007, 0.0002 \\
\bottomrule
\end{tabular}
\end{table}

\begin{figure}[htbp]
\centering
\includegraphics{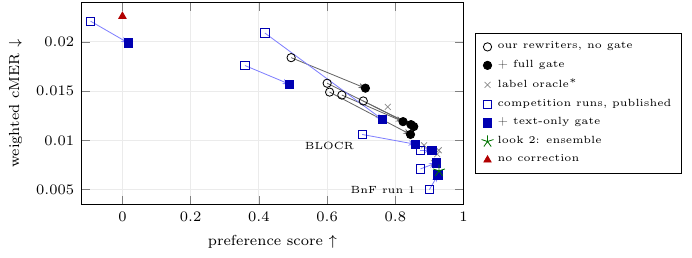}
\caption{Test set: weighted cMER against preference score. Arrows go from the ungated output to the gated output: our five rewriters with the full gate (black) and the published competition runs with the text-only gate (blue), all from look~1. Star: the look-2 ensemble (Sect.~\ref{sec:phase2}). Lower right is better. Values from Tables~\ref{tab:main}, \ref{tab:participants}, and~\ref{tab:look2}.}
\label{fig:pareto}
\end{figure}

\paragraph{Selection protocol and overfitting.} Gates, thresholds, epochs, ensemble pools, and agreement levels were selected on held-out training data, with two exceptions: the look-1 rewriter list and the decision to freeze followed the development results (Sect.~\ref{sec:results}), and the look-2 pool criterion (official cMER of the actual ensemble on the held-out units, replacing the labeled-gain proxy) was adopted after the development scores of the candidate pools had been computed in the same batch; the held-out and development rankings agreed. Before each look the configuration was frozen with SHA-256 checksums (\texttt{test\_hashes.txt}, \texttt{test\_hashes\_p3.txt}, \texttt{test\_hashes\_p4.txt}, released with the code), recomputed and matched before the run. All three looks are reported; ensemble~B was declared the primary system in the look-2 configuration, and look~3 was declared secondary. The held-out units predicted the test ordering of the two ensembles: B and C reach 0.0076 and 0.0073 there and 0.0069 and 0.0066 on test. The look-2 gate was frozen before any Qwen3.8 output existed and lowers the zero-shot Qwen3.8 rewriter's cMER, but only raises the preference score of the LoRA-tuned one. No test number entered any selection, but the direction of the later work was chosen after seeing look~1, and what is built next can follow the test set's quirks even when each configuration is frozen in advance.

\paragraph{Competition runs.} Fig.~\ref{fig:pareto} plots looks~1 and~2 against the published runs, and Table~\ref{tab:look2part} gives the bootstrap comparisons (BLOCR's bootstrap base is 0.0107, against 0.0106 from the scorer's rounded per-cell values). Zakaria-ENSIAS (seven cells, 0.0209) is not in Table~\ref{tab:look2part}; the ensemble's weighted mean over the same seven cells, computed from the scorer's per-cell values, is 0.0066. A $p$ of 0.0002 is the bootstrap floor. Per cell, BnF-Mistral run~1 has the lower cMER on \texttt{icdar2017}-en (+0.0017), \texttt{icdar2017}-fr (+0.0055), \texttt{impresso}-en (+0.0014), \texttt{impresso}-de (+0.0019), \texttt{impresso}-fr (+0.0011, $p=0.13$), and \texttt{dta19-l2} (+0.0035); the ensemble is better on \texttt{dta19-l0} ($-0.0033$) and tied on \texttt{dta19-l1}. The participant files were not gated again in look~2.

\paragraph{Protocol.} Before the run, the checksums of the four adapters, both gates with their rules, and the ensemble configuration (15 files) and seven stored thresholds were recomputed and matched the frozen configuration. The frozen script ran once: the four LoRA rewriters, each without gate, with both gates, and with the label oracle; the ensemble; then, as the pre-declared secondary comparison, zero-shot \mbox{Qwen2.5-72B} and ensemble A. No step was rerun and no harness defect was found; new code was report tooling only. Two, two, six, four, and zero units fell back to the OCR text (empty or truncated output) for \mbox{LoRA-24B}, \mbox{LoRA-32B}, \mbox{LoRA-72B}, \mbox{LoRA-72B-syn}, and zero-shot \mbox{Qwen2.5-72B}, and are scored as unchanged OCR. The zero-shot \mbox{Qwen2.5-72B} rerun with the look-1 prompt and decoding gives 0.0148 against 0.0149 in look~1 ($-0.0002$ [$-0.0003$, $-0.0000$], $p=0.005$): greedy decoding under batched serving is not bit-reproducible across server restarts, at the fourth decimal of the weighted score. Before the look, the development cMER of ensemble B (0.0084) had been extrapolated with the look-1 development-to-test ratios (0.74--0.87) to 0.0062--0.0073 on test, with the explicit prediction that it would not reach BnF-Mistral run~1; the result, 0.0069, is inside that range. The test-best single system (gated LoRA-72B-syn, $-0.0005$ [$-0.0010$, $-0.0001$] against LoRA-72B) was a post-hoc observation and was not acted on. GPU use: 6.41 GPU-hours for the second development phase, 16.57 for the third, and 1.14 for look~2, on two NVIDIA H200 GPUs.

\paragraph{Third look.} \emph{Design.} After look~2, we added \mbox{Qwen3.8-27B} (Apache-2.0; a hybrid model with Gated DeltaNet and full-attention layers), served text-only with thinking off and the look-1 prompt and decoding, zero-shot and with a LoRA adapter (the look-2 recipe; epoch~1 chosen on held-out training units). The look-2 gate was trained and frozen on 30 October 2026; the first Qwen3.8 output was generated on 31 October. It was not retrained or re-thresholded. The protocol, declared before any look-3 number was read, fixed 13 candidate pools with two or three agreement levels and ran an ensemble on test only if the best pool with a Qwen3.8 member had a strictly lower official cMER than ensemble~B on the 360 held-out units; the winner (LoRA-72B, LoRA-72B-syn, both Qwen3.8 rewriters; thresholds 0.667 and 0.123) reached 0.00729 against 0.00758. \emph{Development (reporting only):} zero-shot 0.0156 / 0.729 ungated and 0.0128 / 0.813 gated; LoRA 0.0094 / 0.922 and 0.0089 / 0.976; ensemble~C 0.0082 / 0.975 against 0.0084 / 0.971 for B ($p=0.65$). \emph{Test:} ensemble~C is lower than B in six cells and higher by 0.0001 on \texttt{icdar2017}-en and \texttt{dta19-l1} (neither significant); the gate raises the zero-shot rewriter's cMER on \texttt{icdar2017}-fr (0.0077 to 0.0104) and \texttt{impresso}-en (0.0074 to 0.0085), as in looks~1 and~2. The gated LoRA rewriter is tied with BnF-Mistral run~3 on cMER ($+0.0000$, $p=0.87$; preference $+0.051$, $p=0.003$); ensemble~C is behind run~1 ($+0.0017$ [+0.0012, +0.0021], $p\approx0.0002$) and not significantly ahead of run~3 ($-0.0005$, $p=0.11$). \emph{Protocol:} all frozen checksums and thresholds matched before the run; it was run once, without a harness defect; the ensemble~C members from look~2 were not regenerated. The third look used 3.20 GPU-hours for development and 0.18 for the test look; the project as a whole used 33.83 GPU-hours after the oracle analysis of Sect.~\ref{sec:oracle} (1.08 before it).

\fi

\end{document}